\documentclass{article}

\PassOptionsToPackage{numbers,compress}{natbib}
\usepackage{natbib}

\usepackage[preprint]{neurips_2020}

\usepackage[utf8]{inputenc} 
\usepackage[T1]{fontenc}    
\usepackage{hyperref}       
\usepackage{url}            
\usepackage{booktabs}       
\usepackage{amsfonts}       
\usepackage{nicefrac}       
\usepackage{microtype} 
\usepackage{amsmath}        
\usepackage{graphicx}
\usepackage{wrapfig}
\def\vmu{\vec{\mu}}
\def\vr{\vec{r}}

\DeclareMathOperator*{\argmin}{arg\,min}

\long\def\comment#1{}

\title{Self-Supervised Learning of a \\Biologically-Inspired Visual Texture Model}

\author{%
Nikhil Parthasarathy\\
Center for Neural Science\\
New York University\\
\texttt{np1742@nyu.edu}\\
  \And
  Eero P. Simoncelli \\
  Center for Neural Science, and Courant Inst. of Mathematical Sciences\\
    New York University, and \\
Howard Hughes Medical Institute\\
  \texttt{eero.simoncelli@nyu.edu}
}

\begin{document}

\maketitle
\begin{abstract}
   We develop a model for representing visual texture in a low-dimensional feature space, along with a novel self-supervised learning objective that is used to train it on an unlabeled database of texture images. Inspired by the architecture of primate visual cortex, the model uses a first stage of oriented linear filters (corresponding to cortical area V1), consisting of both rectified units (simple cells) and pooled phase-invariant units (complex cells). These responses are processed by a second stage (analogous to cortical area V2) consisting of convolutional filters followed by half-wave rectification and pooling to generate V2 `complex cell' responses. The second stage filters are trained on a set of unlabeled homogeneous texture images, using a novel contrastive objective that maximizes the distance between the distribution of V2 responses to individual images and the distribution of responses across all images. When evaluated on texture classification, the trained model achieves substantially greater data-efficiency than a variety of deep hierarchical model architectures.  Moreover, we show that the learned model exhibits stronger representational similarity to texture responses of neural populations recorded in primate V2 than pre-trained deep CNNs.  \comment{Additionally, despite the lack of supervision and far fewer learned parameters, the model (when trained on thousands of textures) exhibits classification accuracy competitive with supervised networks pre-trained on millions of natural images.}
\end{abstract}

\section{Introduction}

Most images contain regions of "visual texture" - comprised of repeated elements,  subject to some randomization in their location, size, color, orientation, etc. Humans are adept at recognizing and differentiating materials and objects based on their texture appearance, as well as using systematic variation in texture properties to recover surface shape and depth. At the same time, we are insensitive to the details of any particular texture example - to first approximation, different instances of any given class of texture are perceived as the same, as if they were "cut from the same cloth". This invariance is usually captured through the use of statistical models. Bela Julesz initiated the endeavor to build a statistical characterization of texture, hypothesizing that a texture could be modeled using n-th order joint co-occurrence statistics of image pixels \cite{julesz1962visual}. Subsequent models can be partitioned into three broad categories: 1) orderless pooling of handcrafted raw-pixel features such as local binary patterns \comment{(LBP)} \cite{ojala2002multiresolution, liu2016evaluation}, 2) local statistical models using Markov random fields \comment{(MRFs)} \cite{cross1983markov, chellappa1985classification, derin1987modeling, portilla2000parametric}, and 3)  statistical characterization of fixed convolutional decompositions (i.e. wavelets, Gabor filters, multi-scale pyramids) \cite{bovik1990multichannel, bergen1986visual, heeger1995pyramid, portilla2000parametric, bruna2013invariant, sifre2013rotation}.  More recent models are based on statistics of nonlinear features extracted from pre-trained deep convolutional neural networks (CNN's) \cite{cimpoi2015deep, gatys2015texture, ulyanov2017improved, song2017locally, xue2017differential}. A comprehensive review of these is available in \cite{liu2019bow}. 

The fixed-filter methods are generally chosen to capture features considered fundamental for early visual processing, such as local orientation and scale. Similar filters can be learned using methods such as sparse coding \cite{olshausen1996emergence} or independent components analysis \cite{bell1997independent}. On the other hand, deep learned methods provide great benefits in terms of extracting relevant complex features that are not so easily specified or even described. 

However, recent work in understanding the representation of texture in the primate brain has shown that texture selectivity arises in Area V2 of visual cortex \cite{freeman2013functional, ziemba2016selectivity}, which receives primary input from Area V1. Therefore, it seems that the brain can achieve selectivity for texture in far fewer stages than are commonly used in the deep CNNs. Motivated by this fact, we construct a simple, hybrid texture model that blends the benefits of the aforementioned fixed-filter image decompositions with the power of learned representations. There are two main contributions of our work. First, the model represents textures in a relatively low-dimensional feature space (in contrast to the extremely high-dimensional representations found in CNN models).We propose that this low-dimensional representation can be used to perform texture family discrimination with small amounts of training data when it is coupled with an interpretable non-linear decoder. Moreover, we show that a novel self-supervised learning objective plays an important role in achieving this result. Finally, while pre-trained deep CNNs can achieve better texture classification accuracy, we show that our learned model exhibits much stronger representational similarity to texture responses of real neural populations recorded in primate V2.

\section{Methods}

\subsection{V2Net Model Architecture}
\label{sec:model}
 It is well-known that the primary inputs to V2 are feed-forward outputs from area V1 \cite{girard1989visual, sincich2005circuitry, schiller1977effect}. Inspired by these physiological results, we propose a computational texture model as a two-stage network that functionally mimics the processing in these two early visual areas. 
 
 The V1 stage is implemented using a set of fixed convolutional basis filters that serve as a functional model for V1 receptive fields \cite{ringach2002spatial}. The filters are localized in orientation and scale, specifically utilizing a complex-steerable derivative basis \cite{simoncelli1995steerable, jacobsen2016structured}. We chose a specific set of 4 orientations and 5 scales (octave-spaced) with two phases (even and odd), for a total of $40$ filters. The full set of V1 responses are a combination of both half-wave rectified simple cells and $L_2$-pooled (square root of the sum of squares) complex cells, yielding a total of $60$ feature maps. 
 
\begin{figure}
     \centering
     \includegraphics[scale=0.25]{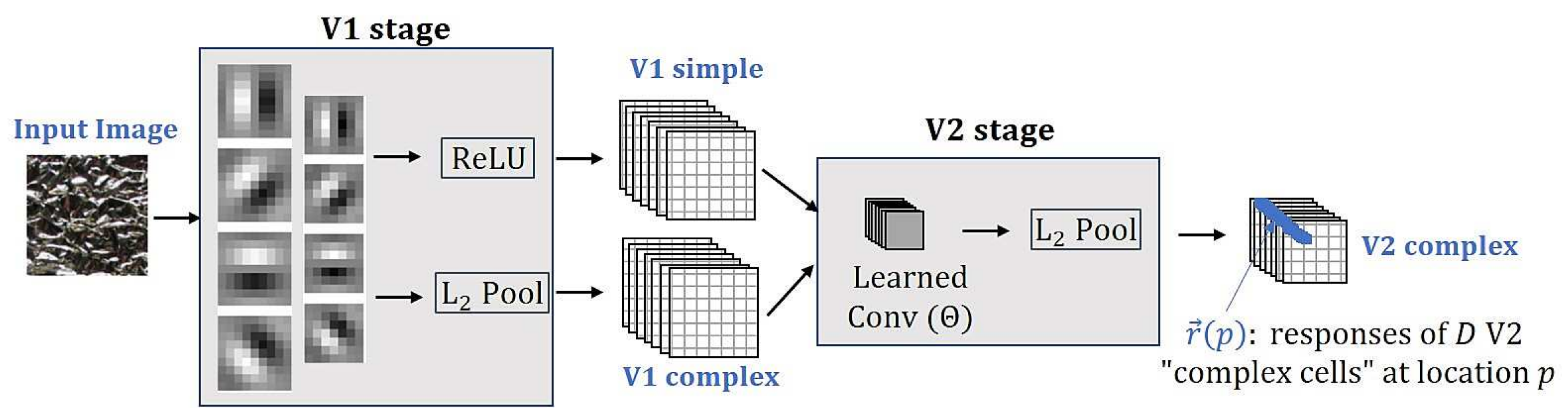}
     \caption{Proposed biologically-inspired texture model architecture. The V1 stage is built using a 5-scale 4-orientation complex steerable pyramid~\cite{portilla2000parametric}, followed by two nonlinearities to generate simple and complex cell responses. The latter uses specialized $L_2$ pooling. The V2 stage consists of convolution with $D$ filters followed by spatial $L_2$ pooling.}
     \vspace{-2\intextsep}
     \label{fig:model-arch}
 \end{figure}

 The V1 responses provide input to a V2 stage that consists of a set of $D$ learned convolutional filters. In the macaque, V1 and V2 are known to have similar cortical surface area and output fibers \cite{wallisch2008structure}, so in our experiments we set D = 60 to match the dimensionality of the V1 and V2 stages of our model. The convolutional layer is then followed by half-wave rectification, spatial $L_2$-pooling and downsampling to produce V2 `complex cell' responses (Fig. \ref{fig:model-arch}). Unlike standard max-pooling, $L_2$-pooling is used in both stages of our model because it is more effective at capturing local energy of responses without introducing aliasing artifacts \cite{bruna2013invariant, henaff2015geodesics}.   
 \subsection{Learning Objective}
 Consider the model in Fig. \ref{fig:model-arch} as a function $f(T ; \Theta)$ that takes as input a texture image $T$, and computes responses based on parameter vector $\Theta = [\Theta_1, ..., \Theta_D]$, which contains the $D$ V2 filters\footnote{each $\Theta_d$ is a 60 x 7 x 7 set of weights, as each V2 filter operates over the full set of 60 V1 channels}. Given a dataset of $N$ texture images ($T_n$) and their corresponding model responses $\vr_n = f(T_n ; \Theta)$, we seek an objective function, $L(\cdot)$, for optimizing the V2 filter weights: $\Theta_{opt} = \argmin_\Theta L( \{f(T_n; \Theta)\})$. We assume a curated image dataset with two properties that underlie the formulation of the objective: 1) individual images contain a single texture type (homogeneous across their spatial extent) and 2) the $N$ images in the dataset represent a diverse set of texture types. 
 
Our learning objective is motivated by the experimental observations in \cite{ziemba2016selectivity} suggesting that V2 represents textures such that responses within texture families (i.e. classes) are largely invariant to variability within the texture families- the responses are less variable within texture families than across families. To learn such a representation, one could simply utilize an objective function that reduces variability of responses to each family while maintaining variability across all families. This can usually be achieved by supervised methods that optimize responses to predict the class identity for an image. However, we desire an objective that has no supervisory knowledge of which images correspond to which texture families. As a result, we propose a contrastive objective that seeks to 1) Minimize the variability of model responses ($\vr_n(p)$) across locations $p$ within each individual texture image and 2) Maximize variability of these responses across neighborhoods sampled from the entire set of $N$  images. Therefore, rather than using labels to enforce grouping of similar texture families, we utilize the natural spatial homogeneity of \textit{individual texture images} as a form of `self-supervision'.   

To formulate this mathematically, we first model the distribution of V2 responses over positions $p$ within each image ($\vr_n(p) \in \mathbb{R}^D$) as multivariate Gaussian, parameterized by the sample mean and covariance: $\vmu_n \in \mathbb{R}^{D}$ and $C_n \in \mathbb{R}^{D \times D}$. The global distribution of responses across all images is then a Gaussian mixture with mean and covariance: $\vmu_g = \frac{1}{N} \sum_{n=1}^{N}{\vmu_n } ~~ ; ~~ C_g = \frac{1}{N} \sum_{n=1}^{N} C_n + (\vmu_n  - \vmu_g)(\vmu_n  - \vmu_g)^\top$. Under this parameterization, the two goals for the objective can be achieved by maximizing the `discriminability' between the individual and global response distributions based on their covariances. A suitable measure of discriminability must capture the differences in both size (total variance) and shape \comment{(dimensionality)} of the distributions. 

There has been extensive work on developing measures that approximate the discriminability between Gaussian distributions based on their mean and/or covariance statistics \cite{bures1969extension, bhattacharyya1946measure, abou2010designing, de2005multimodal, nenadic2007information, dryden2009non, huang2015log}. 
In order to choose a distance for this problem we define a set of criteria the distance must satisfy. First, the distance must be \textit{scale invariant}: global rescaling of the image data should not change the value of the distance measure, which is meant to capture {\em relative} differences in variability. This is especially important for an objective function, as the responses can be arbitrarily scaled by the learned weights. Second, for maximization it is preferable that a distance \textit{have an upper bound} as this can stabilize optimization and avoid degenerate solutions where the distance can take on extremely large, unbounded values. Third, for any given texture image, not all of the V2 dimensions may be important (i.e. the covariance is low-rank), so the distance must be stable in this regime.

Given these criteria, it is clear that many of the statistical distances and manifold-based log-Euclidean distances are problematic because the log transformation is unstable when covariances are low-rank. 
The work of \cite{faraki2016image} has shown that regularizing the log-Euclidean approach with standard covariance shrinkage can lead to large errors, and we have observed this in our experiments as well. A novel attempt to resolve this issue was proposed using a Riemannian optimization method \cite{faraki2016image}, but this method only works for fixed low-rank matrices. As a result, we construct our distance on the form $|| C_1^{1/2} - C_2^{1/2} ||_F$ corresponding to the Bures metric \footnote{Equivalent to the covariance term of the 2-Wasserstein distance between multivariate Gaussian distributions in the special case when the two covariance matrices commute} \cite{bures1969extension, muzellec2018generalizing} . We modify this to make it bounded and scale-invariant, arriving at a novel measure of distance between the global response covariance and that of image $T_n$:
\begin{equation} 
\label{eqn:distance} 
  d_n = \frac{|| ~ C_g^{1/2} - C_n^{1/2} ~ ||_F}{|| ~ C_g^{1/2} ~ ||_F} ,
\end{equation}  
where $(\cdot)^{1/2}$ indicates matrix square-root and $||\cdot||_F$ is the Frobenius norm.
This may be seen as a normalized variant of the log-Euclidean distances \cite{huang2015log}, in which replacement of $\log(\cdot)$ by $(\cdot)^{1/2}$ retains the primary benefit of the log-Euclidean framework (transforming the covariance eigenvalues with a compressive nonlinearity), while remaining stable and well-defined in low-rank conditions.

After calculating the distance in Eqn.~(\ref{eqn:distance}) for each individual image, we then combine over all images to obtain a single scalar objective. To force all distances to be as large as possible, we maximize the minimum of these distances. For stable optimization, we use a soft-minimum function, which yields our variability-based objective:
\begin{equation}
  {\bf L_{\rm var}} = {\rm softmin}(d_1,d_2, \ldots , d_N) = \frac{\sum_n d_n e^{-d_n}}{\sum_n e^{-d_n}} .
    \label{eqn:softmin}
\end{equation}
\comment{Combining all of the above, our variability-based objective is:
\begin{equation}
    {\bf L_{\rm var}} = {\rm softmin}(d_1,d_2, \ldots , d_N, \alpha) .
\end{equation}
}
To allow for robust estimation of the covariance, we make a diagonal approximation where $C_n$ and $C_g$ are each taken to be diagonal. Therefore, the matrix square-roots can be implemented as element-wise square roots of the individual response variances along the diagonal and the Frobenius norm becomes the standard vector $L_2$ norm. However, because a diagonal approximation can be poor if the covariances have strong co-variability, we use an additional orthogonal regularization term to encourage orthogonalization of the V2 filters \cite{bansal2018can}:
\begin{equation}
    {\bf L_{\rm orth}} = ||~ \Theta \Theta^\top - I ~ ||_F .
\end{equation}
Minimizing this loss forces the responses of each channel to be roughly independent and thus more amenable to the diagonal approximation. The final objective is a weighted combination of the two terms:
\begin{equation}
    \label{eqn:finalloss}
    \mathbf{\max_{\Theta}} \left[ \mathbf{L}_{\rm var} - \lambda \mathbf{L}_{\rm orth} \right] .
\end{equation}

\subsection{Evaluation Methodology}
\label{sec:eval}
\begin{wrapfigure}[17]{r}{0.5\linewidth}
    \centering
    \vspace{-\intextsep}
    \includegraphics[width=0.48\textwidth]{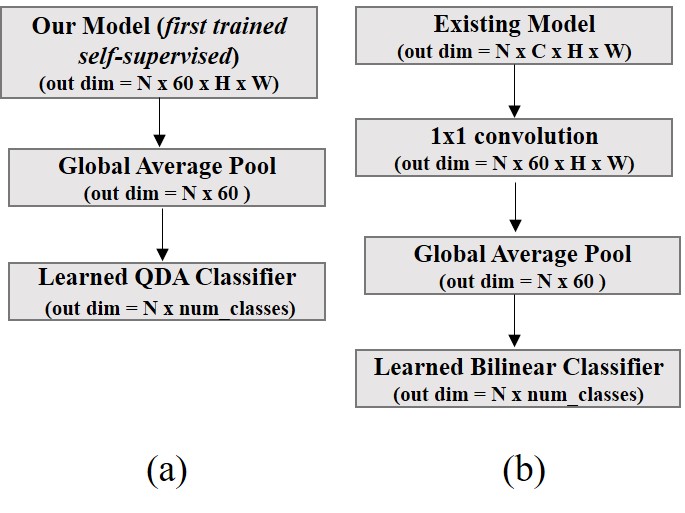}
    \vspace{-\intextsep}
    \caption{(a) Evaluation method for our self-supervised model. (b) Evaluation method for the supervised networks}
    \label{fig:qda}
\end{wrapfigure}

After training the model with the self-supervised objective in Eqn.~(\ref{eqn:finalloss}), we use a separate labeled dataset to train and test a texture family classifier. We first compute the spatially global-average pooled (GAP) responses for each image in the new dataset, such that each image $T_n$ is represented by a single $D$-dimensional vector, $\vmu_n$.  We again make a Gaussian assumption on the distribution of these mean response vectors for each texture family and fit and test a quadratic discriminant classifier (QDA) to predict the texture class labels. This process is shown in Fig. \ref{fig:qda} (a). Although the choice of a QDA classifier is not common, state-of-the-art texture classification methods generally use some form of quadratic feature encoding (Fisher vectors, bilinear layers, etc.) before applying a trained linear classifier (i.e. SVM) \cite{linbilinear, song2017locally, cimpoi2015deep}. Rather than compute all pairwise products, which can be prohibitively expensive in terms of number of parameters, we use mean pooling to produce a low-dimensional representation, followed by a bilinear readout. In our context, a quadratic discriminant is the optimal bilinear method for discrimination under the Gaussian assumption.

One issue with QDA classification is that it requires the estimation of class-covariance matrices. These matrices can only be reasonably estimated when the number of samples per class is much larger than the dimensionality of the features, so QDA is only amenable to low-dimensional feature representations. In our experiments, we compare our model to supervised methods that use popular network architectures as the base feature extractor. However, most of these networks produce very high-dimensional output feature spaces that are usually evaluated with linear classifiers. As a result, we devise an evaluation protocol for these methods such that the dimensionality of the feature representation \textit{and} the expressivity of the classifiers is matched to that of our model (Fig. \ref{fig:qda}(b)). Specifically, we first reduce the dimensionality of the feature representation to match that of our V2Net model (D = 60) using a trained 1x1 convolutional layer (as is common in the literature \cite{howard2017mobilenets, xue2018deep}). This is followed by the same classification procedure as ours: spatial GAP followed by a bilinear classifier. However, because QDA cannot be implemented for supervised, end-to-end trained networks, we use a parameterizable bilinear layer of the form: $x^T A x + B x + c$. The model parameters, 1x1 conv layer, and bilinear are all trained end-to-end, in contrast with our model which is first trained separately with the self-supervised objective. 

One might ask if the dimensionality reduction of the existing network architectures is too restrictive and if our comparisons will be biased because of this. In fact, a similar methodology has shown minimal loss in performance for texture retrieval with PCA reduction down to $64$ dimensions \cite{valente2019comparison}. Therefore, it is unlikely that we are biasing our comparisons by stifling the capacity of the network. Moreover, the 1x1 convolution approach is arguably more effective than PCA because it allows this dimensionality reduction to be optimized in the context of the classification task. Nevertheless, we additionally verified that results for all tested networks were close to those achieved using a linear classifier on the full-dimensional feature space. 

The specific models we compare to are chosen to span a diverse set of methods from the literature:

\textbf{ScatNet}: We implement the front-end two-stage scattering model as described in \cite{bruna2013invariant, sifre2013rotation} that has 5 scales and 8 angles. The scattering model is then fixed and the 1x1 convolution layer and the bilinear classifier are learned. The number of channnels before dimensionality reduction is 681.

\textbf{DAWN(16-init)}: Recent work has performed a similar experiment using a hybrid deep adaptive wavelet network that is found to be more data-efficient than previous methods \cite{rodriguez2020deep}. We implemented the same model and regularization, with 16 initial convolutional layers, followed by the multi-scale representation. The number of channels before dimensionality reduction is 256.

\textbf{ResNet-18}: Based on recent success as a feature extractor for texture recognition \citep{xue2018deep} we also included an 18-layer ResNet model. We extract features from the \textit{layer4} level of the network, as these have been deemed as the most powerful features for texture classification in previous work \cite{xue2018deep, zhai2019deep}. The number of channels before dimensionality reduction is 512.

\textbf{VGG-16}: VGG networks and their variants have been the most common network architectures used for feature extraction in the literature. The work of \cite{cimpoi2014describing, cimpoi2015deep} demonstrated that a Fisher vector decoder, and even linear classification from pooled features of the last convolutional layer, can be effective for texture classification. Based on this work, we used features from the \textit{conv5} layer of a VGG-16 network. The number of channels before dimensionality reduction layer is 512. 

\section{Related Work}
\textbf{Model Architecture.} Many fixed-filter, hierarchical image decompositions have been used in the construction of texture representations that are similar to our V1 stage \cite{bruna2013invariant, simoncelli1995steerable}. However, we note that our V1 responses include both rectified simple cells \textit{and} $L_2$-pooled complex cells. This formulation is motivated by physiological experiments studying the projections of V1 to V2 neurons \cite{el2013visual}, and represents a departure from the classical view of hierarchical visual modeling that assumes only pooled responses are transmitted to the downstream layers \cite{fukushima1980neocognitron,riesenhuber1999hierarchical,bruna2013invariant}. 

Recent deep learning approaches to representing texture have been heavily optimized and hand-crafted for specific tasks such as texture classification \cite{cimpoi2015deep, xue2018deep}, synthesis \cite{ulyanov2017improved, gatys2015texture}, and retrieval \cite{qian2017differential, valente2019comparison}. However, there are a few common themes in these methods that we highlight for their relevance to our model and the models we use for comparison. First, all SoA methods, regardless of task, rely on extraction of features or statistics from deep CNNs trained for object recognition, primarily the VGG and ResNet architectures \cite{simonyan2014very, he2016deep}. With the exception of a few studies \cite{fujieda2018wavelet, rodriguez2020deep}, performing texture classification with networks trained from scratch has been relatively understudied. Second, it has been consistently shown that "orderless" pooling of the features before classification layers results in a far better texture representation. Simple global average pooling (GAP) has been shown to be quite effective \cite{valente2019comparison, xue2018deep, zhang2020uncertainty, dumoulin2016learned} as well as methods that pool based on 2nd-order statistics \cite{cimpoi2015deep, gatys2015texture, linbilinear}.

\textbf{Objective functions.} In the context of texture classification, current human-labeled homogeneous texture databases are few and small, so most deep learning methods transfer features from networks trained with full supervision on an alternative task (typically, object recognition). Some authors have developed limited unsupervised methods based on vector quantization \cite{greenspan1991texture, raghu1997unsupervised}, and non-negative matrix factorization \cite{qin2008unsupervised}. Nevertheless, in concert with CNN models, we believe ours is the first competitive self-supervised learning objective for this problem. 

Conceptually, our objective is inspired by principles of \textit{contrastive learning} that have recently seen much success at competing with more traditional supervised methods \cite{henaff2014local, wu2018unsupervised, oord2018representation, zhuang2019local, henaff2019data}. However, the specific construction of our learning objective differs substantially from these methods as it relies on a diagonal Gaussian parameterization of sample distributions that provides many computational benefits such as easy generalization to incremental learning where the sufficient statistics are updated online without use of large in-memory batches. 
\section{Results}
\subsection{Data-Efficient Texture Classification}

We hypothesized that our objective function enables the learning of a more powerful texture representation from small data. To test this, we used an experimental paradigm similar to \cite{henaff2019data}. We trained and tested all models on varying amounts of data from a texture dataset. We used a modified version of the challenging KTH-TIPS2-b dataset \cite{caputo2005class} for both training and evaluation. The original dataset includes 11 families of textured materials photographed with different viewpoints, illumination levels, and scales. The total dataset is relatively small (4752 images), so we augmented it with 3 rotated versions of each image (90, 180, and 270 degrees) to obtain a total of 19008 samples. As texture representations should be invariant to rotation, this is a sensible augmentation that increases the difficulty of the task. We used the original 4 splits of the KTH-TIPS2-b data (training on 3 splits and testing on the 4th). For all experiments we used a fixed validation set of 3256 images and each test set contained 4752 images. We then conducted three experiments varying the amount of training data (reducing evenly the number of images per texture family). We report results for the full training data (1000 images per family), 50 percent training (500 images per family), and 25 percent (250 images per family).

\begin{wrapfigure}[17]{r}{0.5\textwidth}
    \centering
    \vspace{-2\intextsep}

    \includegraphics[width=0.48\textwidth]{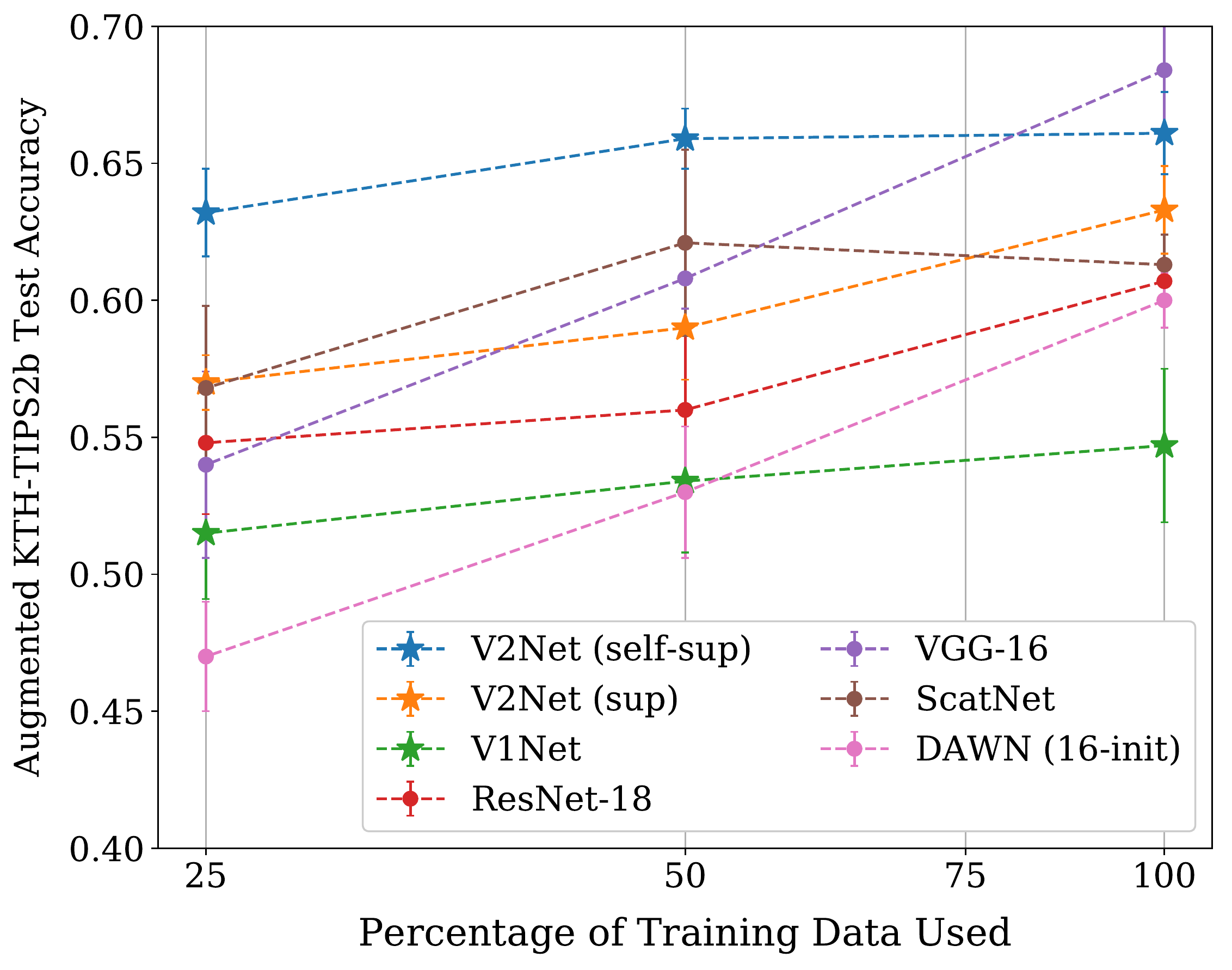}
    \caption{Mean and standard error across the 4 train/test splits as a function of the percentage of training data used.}
    \label{fig:scratchtrain}
\end{wrapfigure}

All models (ours and those listed in Sec. \ref{sec:eval}) were trained from scratch without any pre-trained information. For the supervised networks we varied learning rates (from 0.0001 to 0.01) and batch sizes from (50 to 200) and chose the best model for each train/test split. 
For our model (V2Net (self-sup)), the objective function relies on calculating the global mean and variances over the entire dataset. However, because our training is done through stochastic gradient descent, we approximated these global statistics by the global statistics over batches of 275 images. We chose the batch size heuristically so that individual batch statistics do not deviate significantly from the statistics over the whole dataset. Interestingly, the batch size does not need to be as large as is necessary in most other contrastive learning approaches \cite{henaff2019data, chen2020simple}. We used a learning rate of 0.001 and additionally included a BatchNorm layer at the output of the network to stabilize the global statistics across batches. 

\begin{wrapfigure}[14]{l}{0.42\textwidth}
    \centering
    \vspace{-1.2\intextsep}
    \includegraphics[width=0.4\textwidth]{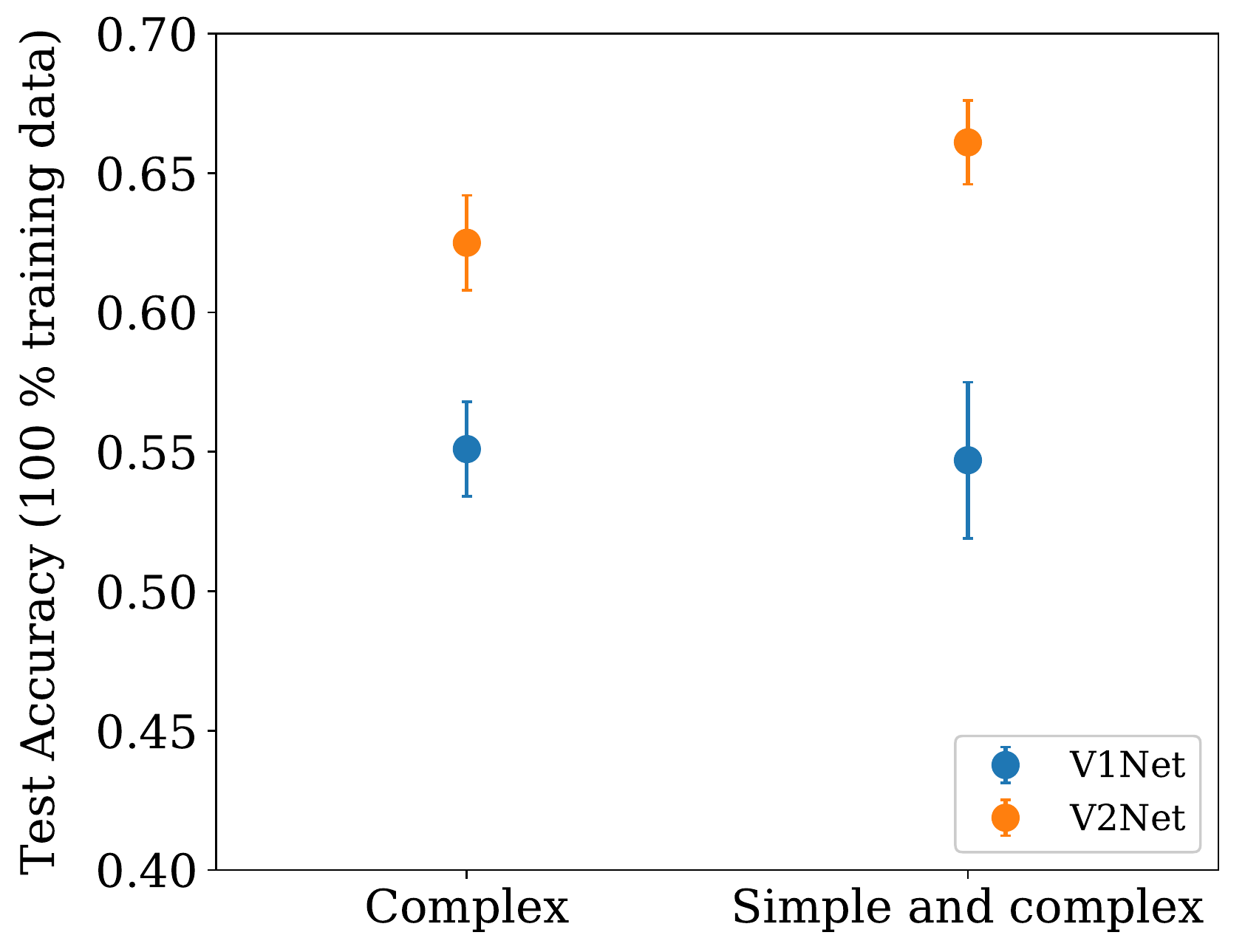}
    \caption{V1 and V2 model comparison based on ablating V1 simple cell contribution.}
    \label{fig:ablate}
\end{wrapfigure}

The results for the 3 training experiments are shown in Fig. \ref{fig:scratchtrain}. We report the mean and standard error for the 4 train/test splits within each experiment. First, we can see that just using the fixed V1 stage (V1Net) followed by QDA provides a reasonable baseline. \comment{This is in fact a reasonable baseline, especially in the 25 \% data experiment (where it is competitive with most of the networks).} This model has marginal performance difference across differing amounts of training data, which can be solely attributed to the estimation error of the class covariances when training the QDA classifier. Second, we find that the  two-stage V2 model performs similarly to the VGG-16 network with full training data, but significantly outperforms all networks when using 50\% or 25\% of the training data, indicating much greater data-efficiency. To better understand the impact of our objective function, we also report results for a network with the same architecture as V2Net, but trained with a supervised cross-entropy loss (V2Net (sup)) \footnote{We use the same bilinear classifier model as was used for the other networks}. As seen in Fig. \ref{fig:scratchtrain}, this network performs comparably to the other supervised networks but still seems to overfit in the small-data regimes. This suggests that even small networks can overfit with small amounts of training data, implying that it is the design of our objective function that allows our network to remain data-efficient in these cases. 

To assess the impact of the inclusion of V1 simple cells in our network, we compared the V2Net classification accuracy to a model trained with V1 simple cells removed. The results are shown in Fig. \ref{fig:ablate}. The performance of both V1 models is roughly the same, and in both cases the V2 model improves on the V1 model. However, the gap between the V1 and V2 performance is noticeably larger when the V1 layer contains both simple and complex cells. This result suggests that a more effective V2 representation can be learned when the inputs come from both simple and complex cells. 

\subsection{Transfer Learning}
\label{sec:transfer}
To verify the generalization of our learning objective, we collected an unlabeled dataset of texture photographs. Original images were manually cropped to be globally homogeneous (by eye) over their entire spatial extent. The scale, viewpoint etc. were not controlled in any particular way, although most textures are on approximately front-parallel surfaces. The types of texture in the dataset span a wide range (including leaves, grass, wood bark, brick, ceramic tile mosaics, etc) that is far more diverse than the KTH-TIPS2-b dataset. We trained our model on 11000 of these images and re-evaluated the performance on the four KTH train/test splits by retraining the QDA classifier. Performance of this pre-trained model slightly improves on the performance of the models trained from scratch (average gain of 1.4 \% mean accuracy across the three experiments) and displays the same level of robustness to the reduction of training data. This demonstrates that our results are not specific to the training dataset and that our learning objective in fact generalizes across texture datasets with very different distributions of images. We additionally compared the performance of our pre-trained (but still self-supervised) network against the ResNet-18 and VGG-16 architectures pre-trained on ImageNet classification. The results of this experiment are given in Appendix \ref{sec:supppre}. Our network does not achieve the performance of these pre-trained networks, but the performance gap (~5-10\%) is surprisingly small given that our model is pre-trained without supervision, using two orders of magnitude fewer images (11k vs. 1M).

\subsection{Selectivity for Natural Texture vs. Spectrally-shaped Noise}
\label{sec:phasescram}
Physiological results in \cite{freeman2013functional, ziemba2016selectivity} suggest that texture selectivity in the brain not only manifests as an ability to separate texture families, but also can also be used to distinguish natural textures from their phase-scrambled counterparts. We constructed a test along these lines to gain a deeper understanding of our learned model and its selectivities. We retrained our V2Net model using phase-scrambled versions of the images from our unlabeled texture dataset from Sec. \ref{sec:transfer}. By training on phase-scrambled images, the model no longer has access to the natural statistics that define textures beyond their spectral power. As a result, if our model is truly capturing higher-order texture statistics, its performance on natural images will drop significantly when trained on the phase-scrambled images. In fact, we find that the average test accuracy of the model trained on phase-scrambled images (V2Net (PS)) is 51.5\% vs. 67.4\% for the model trained on natural images (V2Net (Natural)). Upon further inspection, there are certain texture classes that have high accuracy for the V2Net (PS) model, indicating that these families are readily distinguished using spectral power statistics. We verified that this is also true perceptually: phase-scrambled versions of these classes are visually similar to the original images. However, the classes where there is a large deviation between V2Net (Natural) and V2Net (PS) are those where the phase-scrambled images carry little information about the original texture. For more details, see Appendix \ref{sec:suppps}.

\subsection{Texture Representational Similarity}
Having established that our learned texture model reproduces the qualitative texture selectivities seen in populations of V2 neurons, we explored this relationship quantitatively by comparing the representational similarity between our model and recorded responses of V2 neurons to texture images. We used the dataset described in \cite{freeman2013functional, ziemba2016selectivity}, which provides electrophysiological recordings of 103 V2 neurons responding to 15 samples of textures from 15 different texture families. To understand the representational similarity between our model and the neural data at the level of texture families, we first computed the averaged response (across samples) of both the model and neural responses to each of the 15 texture families. Next, for each representation, we constructed a dissimilarity matrix based on the pairwise correlation distance. There are many distances one could choose but the correlation distance is one of the most common and performs fairly robustly in comparison with distances such as euclidean distances  \cite{mehrer2020individual, kriegeskorte2008representational}. As has been noted in the literature \cite{nili2014toolbox}, it is not common to assume a linear relationship between dissimilarity matrices, but it is rather more appropriate to assume the model RDM predicts the rank order of the dissimilarities \cite{nili2014toolbox}. Therefore, we computed the Spearman rank correlation between the dissimilarity matrices of our model and the V2 neural data. We repeated this process for our V2Net model that uses only V1 complex cells (V2Net (Comp)). Finally, we performed the same analysis for all of the major layers from pre-trained ResNet-18 and pre-trained VGG-16 networks, reporting results for the layer with the best correlation. The results are summarized in Table 1. For more details on the physiology data and image presentation see Appendix \ref{sec:suppneural}.

We find that both V2Net representations are significantly more correlated with the V2 population representation than either deep CNN, and that the inclusion of simple cells again offers a noticeable improvement. Additionally, for both of the pre-trained deep networks we find that the best performance occurs in early layers (layer1 of ResNet and block1pool of VGG). These layers are arguably at the right level in the visual hierarchy (2 or 3 convolutional layers deep) to be matched to V2, but we note that previous work has suggested that texture classification improves when taking features from deeper layers (e.g., VGG conv5) \cite{cimpoi2015deep}. We hypothesize that this is a result of the supervised learning objective used to train these networks. Because of this, the pre-trained deep networks are able to achieve higher overall texture classification accuracy than our current model, but are not able to capture the physiology as well. This suggests that stacking a hierarchical model on top of our learned network may lead to an improvement in SoA classification performance while maintaining consistency with biological architectures.
\begin{table}
\begin{center}
 \begin{tabular}{||c | c | c | c | c||} 
 \hline
  & V2Net & V2Net (Comp) & Res (layer1) & VGG (block1pool) \\ [0.5ex] 
 \hline
 Spearman Corr. & \textbf{0.658} & 0.597 & 0.410 & 0.405 \\ [1ex]
 \hline
 \end{tabular}
 
 \end{center}
 \caption{Spearman correlation between model RDMs and the V2 neural data RDM representing texture families. V2Net (Comp) refers to the V2Net model trained with only V1 complex cells.}
 \vspace{-1\intextsep}
 \end{table}
 
\section{Discussion}
In this work, we demonstrate successful data-efficient self-supervised learning of a simple, yet powerful computational model for representing texture. Rather than learn a very high-dimensional representation followed by linear classification, we use a simpler two-stage model whose responses are then decoded with an \textit{interpretable} non-linear decoder (QDA). This provides the benefit that moving forward we can more easily probe the underlying learned feature space and understand explicitly how those features impact decoding of texture families (through their covariance structure). In fact, we are not the first to propose such a scheme in the context of neural decoding as QDA has been shown to provide a possible basis for a biologically-plausible non-linear decoding method that can explain quadratic transformations that have been observed between layers of processing in the visual system \cite{pagan2016neural, yang2020revealing}. Within this framework, we show that a modification of the common view of hierarchical visual processing (reminiscent of skip-connections \cite{he2016deep}), that includes both V1 simple and complex cells as input to a second V2-like processing stage can provide functional benefits in the learning of the texture representation both in terms of classification accuracy and representation similarity with recording neurons in primate area V2. More importantly, we demonstrate that smaller networks do not necessarily perform much better with small training data, but that learning robustly from small numbers of training examples required the development of a novel self-supervised learning objective.  

Our learning objective is inspired by recent unsupervised contrastive objectives (separating positive examples from a collection of negatives) \cite{zhuang2019local, oord2018representation, henaff2019data, wu2018unsupervised}. While these methods are general, in they are non-parametric with respect to the distribution of the data, we believe that our parameterization in terms of mean and covariance allows our method to 1) constrain learning in small data regimes and 2) provide opportunities to explore more biologically plausible on-line learning implementations. In particular, it is implausible that the brain can store all samples of the global distribution, and our parameterization allows for on-line sequential update of the mean and covariance statistics for each observed image. 

Finally, our model currently assumes a dataset of homogeneous textures as input, enabling a simple form of objective that minimizes spatial variability of the responses across each image. We are currently extending this to allow learning from whole natural scenes, by minimizing variability of responses within {\em local} spatial neighborhoods, while maximizing global variability. This is motivated by the local consistency of natural images - nearby spatial regions are more likely to be similar than distant ones. In fact, there have been some efforts to use spatial coherence as a learning signal \cite{becker1995spatial, ji2018invariant, danon2019unsupervised, jean2019tile2vec}, splitting the image into independent patches that are processed as inputs to the model during learning. Our objective offers an alternate methodology that can process full images while imposing the locality constraint in the response space. Because of the layer-wise nature of our objective, there is also the potential to extend the method to learn filters in multiple stages of a hierarchical model.

\renewcommand\refname{\vskip -1cm}

\section*{Broader Impact}
The interplay between machine learning and neuroscience is something that has helped progress both fields throughout the previous decades. Our work lies at the intersection of these two fields and aims to provide new accounts of how the visual system processes information by utilizing computational methods and constraints from physiology.  In the context of machine learning, achieving generalizable unsupervised learning is widely considered to be one of the most important open problems, and our work provides a novel unsupervised method for learning representations throughout a hierarchical model, that is additionally more biologically plausible due to its self-supervised and layer-wise nature. Our work also provides insight into understanding how the primate visual system encodes natural images. Much of the field of visual neuroscience has focused on building models of either early sensory areas through primary visual cortex, or late-sensory areas (i.e V4/IT) that can be to some extent captured by deep network models. Computational models of mid-visual areas are lacking in comparison. The model described here aims to bridge this gap, both from a perspective of understanding the feature representation in area V2, but also providing physiologists with stronger hypotheses and experimental design constraints for probing mid-visual areas.

In terms of social and ethical implications, the high-level goal of generalizable unsupervised learning has the risk of creating uninterpretable AI models that will be used as replacements for more interpretable human-level decision-making. It is generally thought that lower dimensional models such as ours can allow for greater interpretability and analysis. Finally, to the extent that it provides a good description of primate mid-visual processing, our model offers potentially important societal value in the context of medical applications such as human visual prosthesis development. \comment{Visual prosthesis research will benefit substantially from detailed models of how prosthetic signals fed to early sensory areas are transformed through the visual hierarchy. }

\section*{Acknowledgments}
This work has been supported by the Howard Hughes Medical Institute (EPS and NP) and the NIH Training Grant in Visual Neuroscience (T32 EY007136-27)
\section*{References}
\comment{References follow the acknowledgments. Use unnumbered first-level heading for
the references. Any choice of citation style is acceptable as long as you are
consistent. It is permissible to reduce the font size to \verb+small+ (9 point)
when listing the references.
{\bf Note that the Reference section does not count towards the eight pages of content that are allowed.}
}

{\small
  \bibliographystyle{unsrtnat}
  \bibliography{egbib}
}

\normalsize
\newpage
\appendix

\section*{Appendix}
\label{sec:suppmaterials}
\section{Transfer Learning with Pre-trained Networks}
\label{sec:supppre}

In addition to comparing networks trained from scratch on our modified KTH dataset, we also tested the performance of features transferred from pre-trained versions of our V2Net model, VGG-16, and ResNet-18. We pre-trained our model on a dataset of 11000 unlabeled image patches using our self-supervised objective. Example images from this dataset are provided in Fig. \ref{fig:tex-ims}:
\begin{figure}[ht!]
\centering
    \includegraphics[scale=0.4]{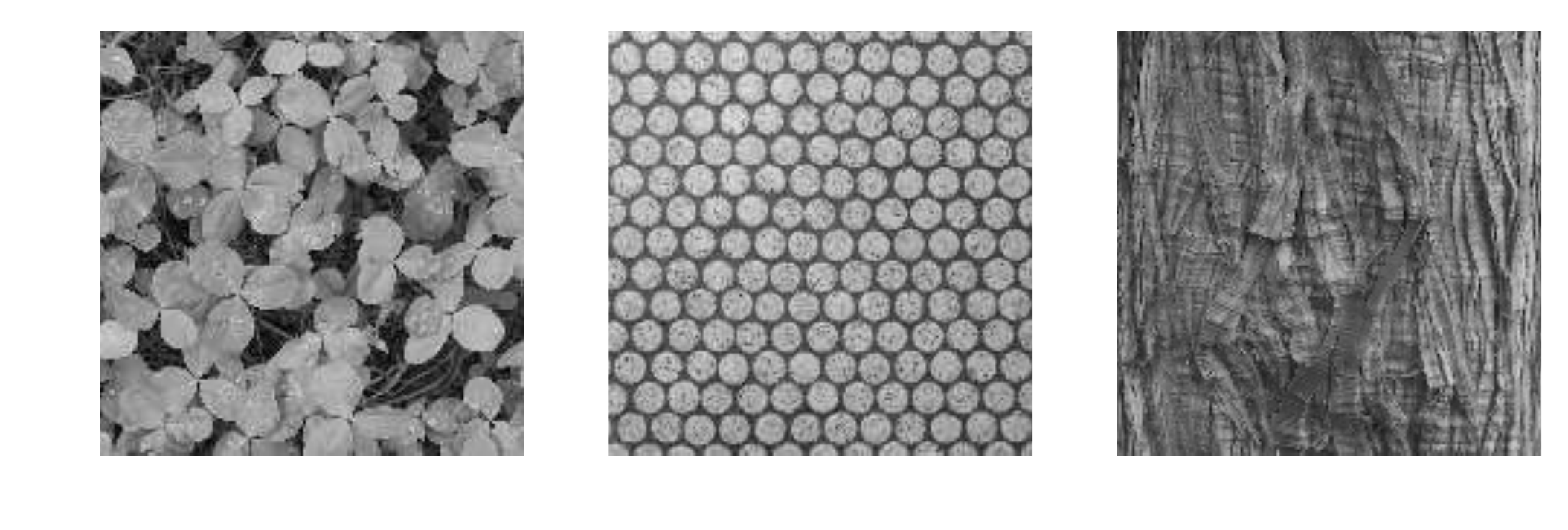}
    \caption{Example texture images from our hand-curated dataset, comprised of a large collection of natural textures that are unlabelled, but diverse in content and homogeneous across their spatial extent.}
    \label{fig:tex-ims}
\end{figure}

The VGG and ResNet networks are pre-trained on the supervised task of object recognition using 1 million images from the ImageNet database. 

\begin{figure}[ht!]
    \centering
    \includegraphics[scale=0.4]{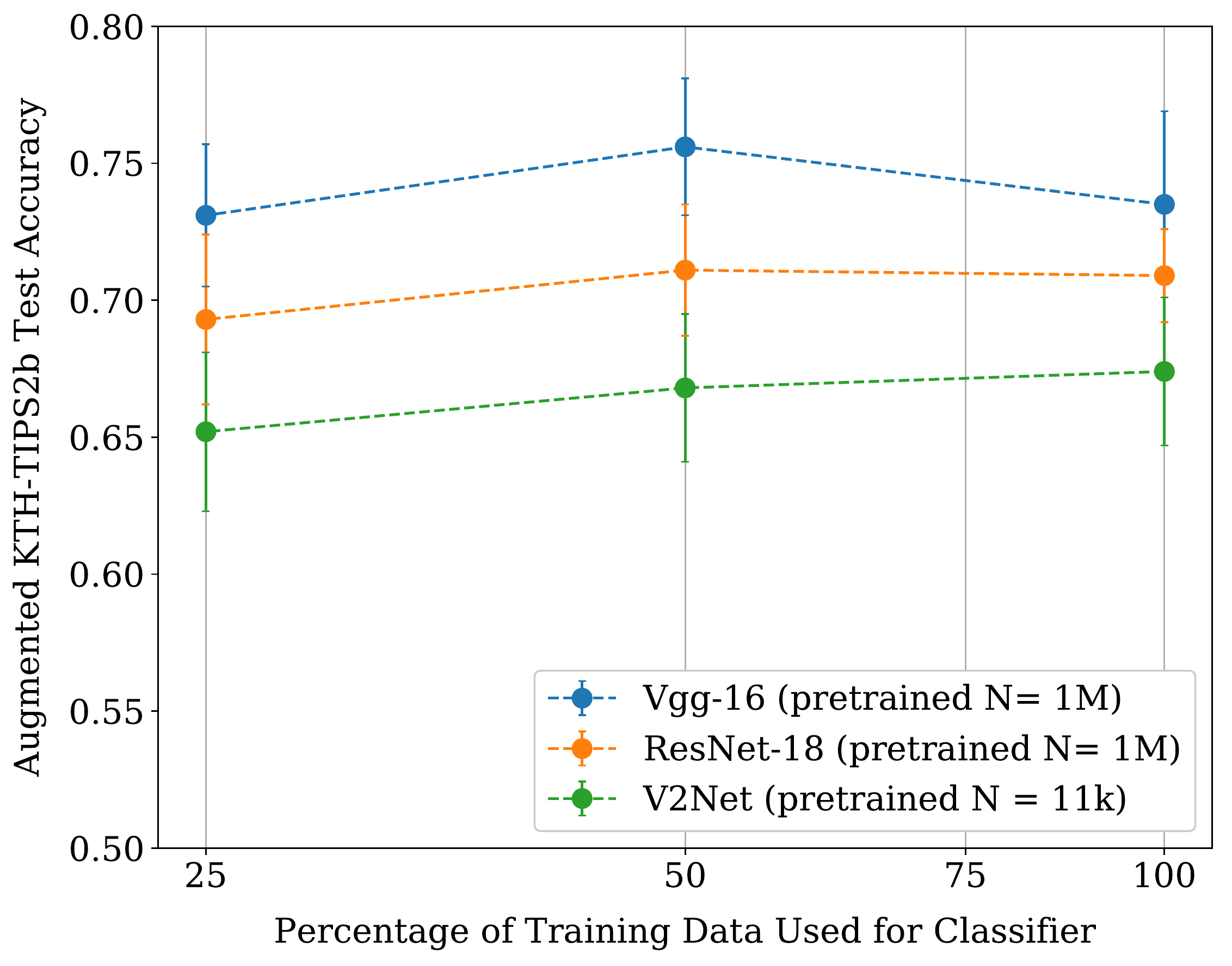}
    \caption{Mean and standard error computed across the 4 train/test splits on our KTH dataset (for each experiment where 25 \%, 50 \% and 100 \% of training data is used to train the classifier weights). N refers to the number of images used to pre-train each model.}
    \label{fig:pretrainplot}
\end{figure}

We used these pre-trained networks as feature extractors, and retrained the respective classifiers (See Fig. \ref{fig:qda}) for texture classification. Results are shown in Fig. \ref{fig:pretrainplot}.

For all of the models, we find the classifier does not require large amounts of training data - performance is relatively constant across the different amounts of training data used. For the full (100\%) classifier training set, our model achieves 67\% -  the performance gap (~5-10\%) relative to the pre-trained CNNs is surprisingly small given that our model is pre-trained without supervision, using two orders of magnitude fewer images (11k vs. 1M). 

\section{Selectivity for Natural Texture vs. Spectrally-shaped Noise}
\label{sec:suppps}
\begin{figure}[ht!]
    \centering
    \includegraphics[scale=0.45]{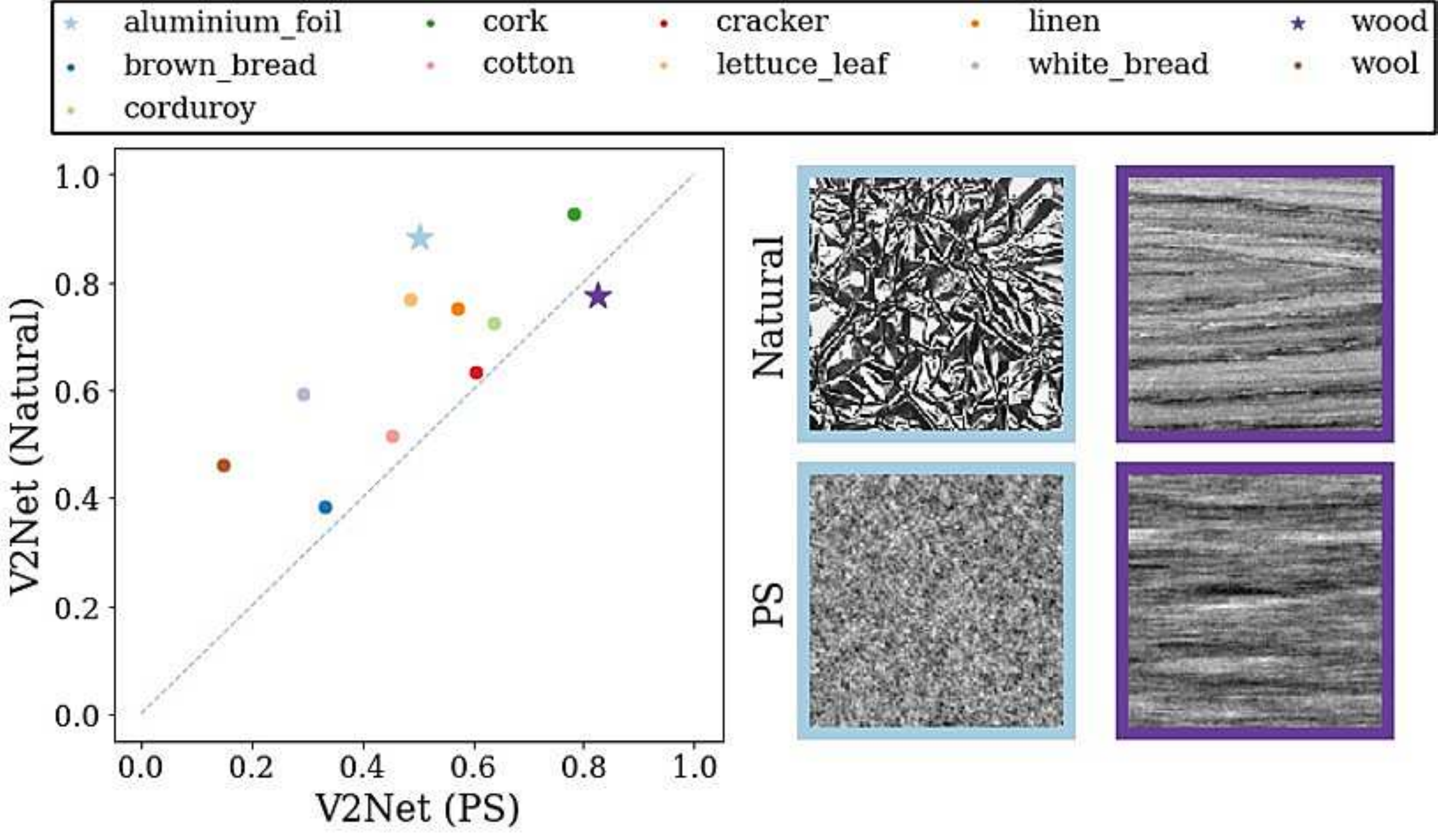}
    \caption{Left: Average KTH test accuracy (averaged over 4 splits with 100 \% training data) for the V2Net (Natural) vs. V2Net (PS) models. Right: example images (both natural and phase-scrambled) for two texture families. For the `aluminum foil' family, the phase-scrambled image removes the higher-order content that is necessary for identifying the texture. For the `wood' family, the phase-scrambling does not alter perception of the texture significantly, because its appearance is primarily determined by spectral content.}
    \label{fig:psplot}
\end{figure}

For each of the 11 texture families in the test dataset, we plot the mean accuracy our model trained on natural images (V2Net (Natural) vs. our model trained on phase-scrambled images V2Net (PS)). Fig. \ref{fig:psplot} shows that the model trained on natural images performs better for most texture families, since it is able to capture higher-order natural statistics. If we visualize an example of one of these classes (aluminum foil), we see that this is because the scrambling of phase destroys content that is critical in defining that texture. However, for a few families, the performance of the V2Net (PS) model is about the same as the V2Net (Natural) model because certain texture families (e.g. wood) are primarily defined by their spectral content (and thus not altered significantly by phase-scrambling).

\section{Representational Similarity Methods and Data}
\label{sec:suppneural}
Here, we provide more details about the dataset and methods used for the representational similarity analysis presented in the main text. The neural data taken from \cite{ziemba2016selectivity} consists of electrophysiolgical recordings of 103 V2 neurons from anesthetized adult macaque monkeys. As is done in the original analysis, we averaged spike counts within 100-ms time windows aligned to the response onset for each single unit. To gaussianize the neural responses, we applied a variance-stabilizing transformation to the spike counts for each neuron ($r_{gauss} =  \sqrt{r_{poiss}} + \sqrt{r_{poiss} + 1}$). 

The visual stimuli used in the experiment are synthetic texture stimuli generated using the procedure described in \cite{portilla2000parametric}. A set of 15 grayscale texture photographs are used as the examples for 15 different texture families. From these seed images, 15 samples are generated for each family to provide sample variation across the family. The original stimuli have a size of 320 x 320 pixels and are presented to every V2 unit at a size of $4^{\circ}$, within a raised cosine aperture (this window was larger than all of the receptive fields of the neurons at the recorded eccentricties). For our representational similarity experiments, we thus pre-processed the images for input to the models such that they are resized to the appropriate pixel dimensions (224 x 224) and presented within a $4^{\circ}$ raised cosine aperture. This ensures that the models receive inputs with content comparable to what is seen by the neurons in the V2 recordings. 
\end{document}